\documentclass[letterpaper, 10 pt, conference]{ieeeconf}  

\IEEEoverridecommandlockouts                              

\usepackage{multirow}
\usepackage{amsmath} 
\usepackage{amssymb}  

\usepackage{makecell}

\usepackage{graphicx}
\usepackage{algorithm}
\usepackage{algpseudocode}
\algnewcommand\algorithmicswitch{\textbf{switch}}
\algnewcommand\algorithmiccase{\textbf{case}}
\algnewcommand\algorithmicassert{\texttt{assert}}
\algnewcommand\Assert[1]{\State \algorithmicassert(#1)}%
\algdef{SE}[SWITCH]{Switch}{EndSwitch}[1]{\algorithmicswitch\ #1\ \algorithmicdo}{\algorithmicend\ \algorithmicswitch}%
\algdef{SE}[CASE]{Case}{EndCase}[1]{\algorithmiccase\ #1}{\algorithmicend\ \algorithmiccase}%
\algtext*{EndSwitch}%
\algtext*{EndCase}%

\usepackage{environ}
\NewEnviron{resizealign}{\sbox0{
    $\begin{matrix}\displaystyle\BODY\end{matrix}$}%
  \sbox1{$(\theequation)$}%
  \sbox2{\parbox{\dimexpr \wd0 + 2\wd1}%
    {\begin{align}\BODY\end{align}}}
  \noindent\resizebox{\columnwidth}{!}{\usebox2}%
}

\title{\LARGE \bf
Temporal and Physical Reasoning for Perception-Based Robotic Manipulation*
}

\author{Felix Jonathan, Chris Paxton, and Gregory Hager
\thanks{This work was funded by NSF NRI Grant Award No. 1637949.}
\thanks{Felix Jonathan, Chris Paxton, and Gregory Hager are with the Department of Computer Science, Johns Hopkins University, 3400 N. Charles St. Baltimore, MD 21218-2686, USA.
        {\tt\small fjonath1@jhu.edu, cpaxton@jhu.edu, hager@cs.jhu.edu}}%
}

\begin{document}

\maketitle
\thispagestyle{empty}
\pagestyle{empty}

\begin{abstract}
Accurate knowledge of object poses is crucial to successful robotic manipulation tasks, and yet most current approaches only work in laboratory settings. Noisy sensors and cluttered scenes interfere with accurate pose recognition, which is problematic especially when performing complex tasks involving object interactions.
This is because most pose estimation algorithms focus only on estimating objects from a single frame, which means they lack continuity between frames.
Further, they often do not consider resulting physical properties of the predicted scene such as intersecting objects or objects in unstable positions.
In this work, we 
enhance the accuracy and stability of estimated poses for a whole scene by enforcing these physical constraints over time through the integration of a physics simulation. This allows us to accurately determine relationships between objects for a construction task.
Scene parsing performance was evaluated on both simulated and real-world data. We apply our method to a real-world block stacking task, where the robot must build a tall tower of colored blocks.
\end{abstract}

\section{Introduction}

Despite the increasing popularity of robot manipulators in various industries, autonomous robots which manipulate objects in an unconstrained environment are still uncommon, since estimating object positions in diverse and cluttered scenes have low reliability.
Accurate manipulation requires object pose estimates that both minimize error and
conform with properties of the physical world. No rigid objects in the scene will ever intersect with each other; all static objects will have a stable position.
Given noisy sensor data, it is possible that supposedly ``perfect'' object poses from a pose estimate will be inconsistent with physical characteristics of a scene. 
In other words, the most likely pose for an object given sensor data is not necessarily the most likely pose for an entire scene.
Our proposed method uses this simple fact to reject impossible scenes and dramatically improve robot performance on complex manipulation tasks.

Vision-based pose estimation algorithms often intelligently perform RANSAC-like sampling of poses for known object models on RGB-D data~\cite{aldoma2012global, papazov_burschka_2011, papazov_haddadin_parusel_krieger_burschka_2012}. While these demonstrate reasonable pose estimation accuracy, they do not impose these physical constraints on resulting estimates.
These limitations may prevent robots from performing tasks that require consistent pose assignment to compute accurate information about object relationships, such as structure assembly tasks.
Since the pose estimation process does not enforce any temporal and physical constraints, the estimated poses are subject to random pose fluctuations between detections and may contain some physical violations such as intersecting objects.


\begin{figure}[tb]
  \centering
  \includegraphics[width=\columnwidth]{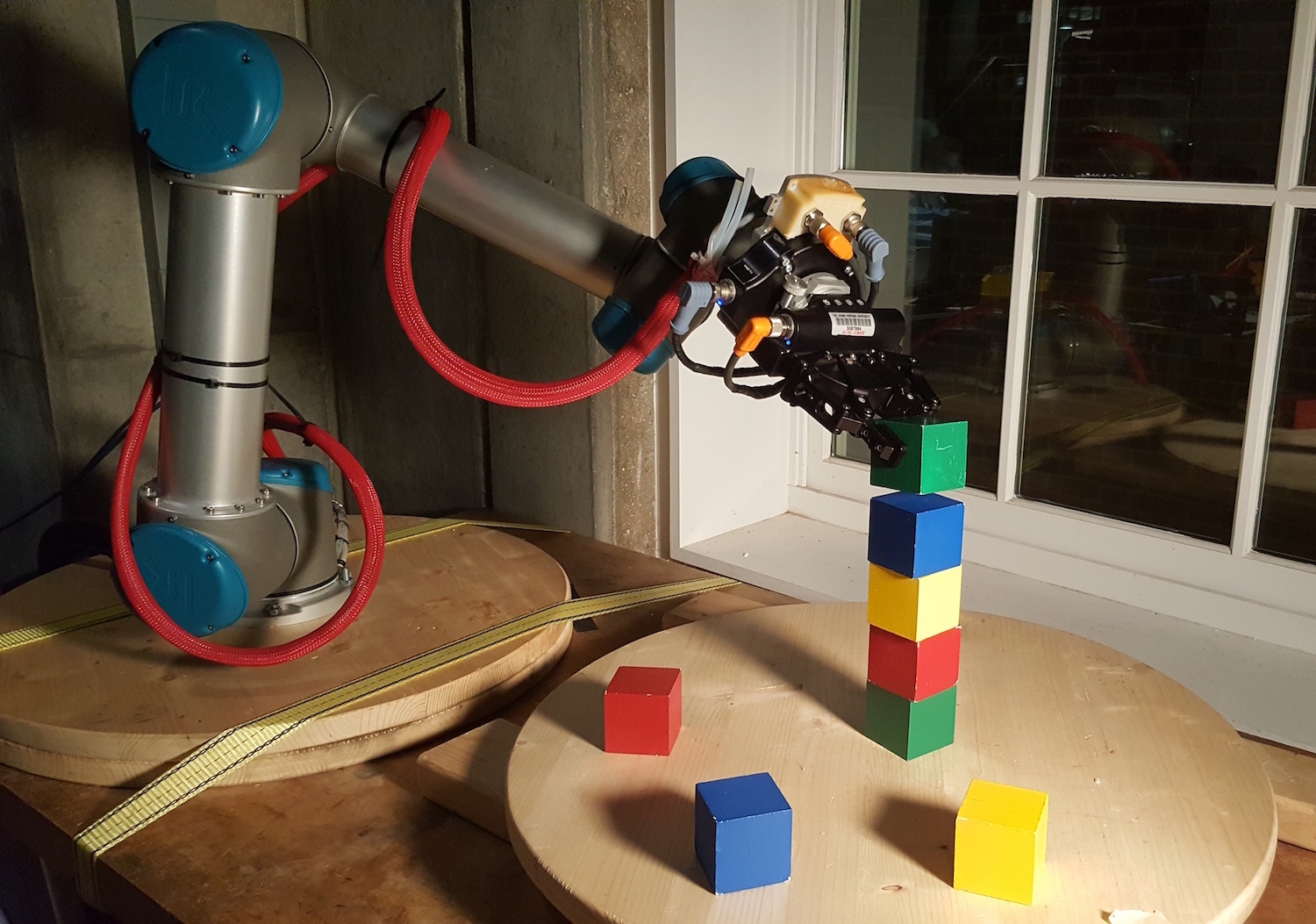}
  \caption{A sample picture from the case study.}
  \label{fig:robot_initial_setup}
\end{figure}

Previous work in Sequential Scene Parsing (SSP) addresses these limitations by iteratively updating a representation of the world while maintaining physical consistency~\cite{hager_wegbreit_2011, brucker_leonard_bodenmuller_hager_2012}.
However, these approaches have serious limitations of their own.
Work by Hager et al. is only applicable to geometric primitives~\cite{hager_wegbreit_2011}.
Brucker et al. used an approach based on SURF features, which only work on textured objects. They also report pose error of up to several centimeters, which is insufficient for accurate robotic manipulation.


We present a new algorithm for SSP that is based on semantic segmentation and 3D model-based pose estimation. This algorithm uses a physics engine to approximate the new scene by incorporating knowledge from previous scene, segmented point cloud, and the object hypotheses generated by pose estimators.
Our approach approximates the optimal scene with linear complexity to the total number hypotheses from detected objects.
For each object hypothesis, the physics simulation generates a more physically consistent configuration while also maintaining its consistency with the data.
Objects are iteratively added to the scene to construct an accurate estimate taking into account physical constraints, similarity to data, and temporal consistency. 
The configuration that maximizes the scene probability is then used to represent the previous scene for the next detection. 


Our contributions are:
\begin {itemize}
\item A simulation-based approach for computing scene quality and consistency. 
\item A method for efficient scene evaluation when computing the optimal scene.
\item Experimental validation on simulated and real-world data, and on a real-world robotic structure assembly task. 
\end {itemize}

\section{Related Work}
Since the arrival of affordable RGB-D cameras, interest in scene modeling has grown substantially.
KinectFusion provides a method to reconstruct the scene to a 3D model using a SLAM based method and physics interaction~\cite{newcombe2011kinectfusion}.
This work was continued in DynamicFusion \cite{newcombe2015dynamicfusion}, which added support for moving and deformable objects.
Point-based Fusion supports building a nonstatic scene~\cite{keller2013real} and ElasticFusion can generate a room-sized model~\cite{whelan2015elasticfusion}. These scene building tools can be used to improve object segmentation results, such as in the work of Tateno, et al.~\cite{tateno2015real}, who use Point-based Fusion \cite{keller2013real} as an input to real-time unsupervised object segmentation.

Machine learning approaches perform scene construction together with completion and semantic segmentation~\cite{song2016semantic,zheng2013beyond,jia20133d,silberman2012indoor}.
In work by Song, et al. \cite{song2016semantic}, the scene is both labeled and volumetric occupancy is predicted to generate a voxelized representation of the scene. Similar work has been done by Zheng, et al. \cite{zheng2013beyond} utilized geometric and physical reasoning for semantic scene completion. Jia et al. \cite{jia20133d} generate a segmented scene by using bounding boxes for jointly optimizing physical stability and segmentation of each object.
Silberman, et al. \cite{silberman2012indoor} use physical reasoning in the form of support inference to improve scene segmentation on a cluttered indoor room. 
However, these works do not look at estimating accurate object poses for manipulation.

Previous work such as ObjRecRANSAC \cite{papazov_haddadin_parusel_krieger_burschka_2012} and Global Hypothesis Verification \cite{aldoma2012global}, use point pair correspondences between models and the scene to generate object pose hypotheses. These then perform ICP to minimize estimated pose accuracy. However, these approaches do not enforce compliance with physical constraints.

Other work has evaluated object poses based on their physical world compliances. Wagle, et al. tested collisions and stabilities of object hypotheses in a physics engine to prune invalid object hypotheses~\cite{wagle2010multiple}, although it only supports geometric primitives and does not use the temporal information of the scene to maintain object's temporal consistency. Li, et al. propose a machine learning approach that uses a model trained on synthetic data generated from physics simulation to estimate the current scene stability~\cite{li2016fall}, which can potentially be used to achieve a faster scene evaluation than our simulation-based evaluation.

Li, et. al. use unsupervised segmentation as an input to train and performs object classification and RANSAC-based pose estimation from an unsupervised object segmentation result~\cite{li2016incremental}. This research shows an improved performance in both object classifications and poses estimations compared to previous work~\cite{li2015beyond, li2015bridging} given that the input data give a complete representation of each object on the scene. This work is capable of generating good pose estimates for individual objects and has some incremental scene building ability, but does not enforce physical constraints.

\section{Scene Probability Formulation}
The goal of performing scene parsing is to generate a scene model $s \in S_C$ which best explains the sensor data $D$ while complying with physical constraints, where $S_C$ is the set of physically consistent scene configurations. This can be posed as the problem of maximizing $\Pr(s|D)$ subject to these constraints.



Let $t$ denote the current frame and $s_t$ denote the current scene. The set of prior scenes is given by $S_{t-1}$. Assuming that the sensor pose is static and the scene dynamics are Markov, then based on the law of total probability, $\Pr(s|D)$ is given by:
\begin{equation} \label{eq:scene_probability_base}
\Pr(s_t|D) = \int_{S_{t-1}} \Pr(s_t|D,S_{t-1}) \Pr(S_{t-1}) dS_{t-1}
\end{equation}

Applying Bayes' theorem and simplifying yields:
\begin{resizealign} \label{eq:scene_probability}
\Pr(s_t|D) \propto \int_{S_{t-1}} \Pr(D|s_t) \Pr(s_t|S_{t-1}) \Pr(S_{t-1}) dS_{t-1}
\end{resizealign}

We use the prior MAP estimate instead of the entire prior scene distribution to simplify the scene estimation. This allows us to use only a single prior scene $s^*_{t-1}$ to compute the probability of the best scene $s^*_t$:
\begin{equation} \label{eq:scene_probability_unsimplified}
\Pr(s^*_t|D) \propto \max_{s_t} \Pr(D|s_t) \Pr(s_t|s^*_{t-1})
\end{equation}

We now
approximate scene physical consistency by introducing an explicit physical consistency measure $C$ with an associated probability that penalizes physical consistency violations. We independently observe sensor measurements $D$ and physical consistency $C.$ This allows us to replace the maximization of $\Pr(s^*_t|D)$ with $\Pr(s^*_t|C,D)$:
\begin{align}
\Pr(s^*_t|D) &\propto \max_{s_t} \Pr(D,C|s_t) \Pr(s_t|s^*_{t-1}) \nonumber \\
    &\propto \max_{s_t} \Pr(D|s_t) \Pr(C|s_t) \label{eq:scene_probability_simplified}\Pr(s_t|s^*_{t-1}) 
\end{align}
This lets us compute physical consistency explicitly and introduce it into the optimization. Let $O_t=[o_{t,1},o_{t,2},o_{t,3},...,o_{t,N}]$ be all objects in the scene and $\Pr(C(o_{t,i})|s_t)$ be the measured physical consistencies of an object $o \in s_t$. The consistency of the scene, $\Pr(C|s_t)$, is given by $\Pr(C|s_t) = \prod\limits_{i=1}^N \Pr(C(o_{t,i})|s_t)$. We consider 3 constraints in $C = \{\mathcal{B},\mathcal{C},\mathcal{S}\}$:
\begin{itemize}
\item Stability ($\mathcal{B}$): Objects are static when no external force is applied. Stability measures the temporal changes of an object under the effect of gravity.
\item Collision ($\mathcal{C}$): Objects cannot have overlapping volumes; therefore, any collisions will generate penalties to the physical consistency.
\item Support contribution ($\mathcal{S}$): Objects always have consistent interaction with each other from one scene to the next, which can be represented with object support relationships.
\end{itemize}

All $C$ components can be estimated by a physical simulation. Substituting the physical consistency of each object with its components yields:
\begin{resizealign} \label{eq:physical_consistency_probability}
\Pr(C(O_t)|s_t) = \prod\limits_{i=1}^N \Pr(\mathcal{B}(o_{t,i})|s_t) \Pr(\mathcal{C}(o_{t,i})|s_t) \Pr(\mathcal{S}(o_{t,i})|s_t)
\end{resizealign}

The scene transition $\Pr(s_t|s^*_{t-1})$, similar to stability, can be generated by penalizing the movement for all objects in $O_t$ from the previous scene to the current scene. This can be calculated with:
\begin{equation} \label{eq:scene_transition_probability}
\Pr(s_t|s^*_{t-1}) = \prod\limits_{i=1}^N \Pr(o_{t,i}|o_{t-1,i})
\end{equation}

\subsection{Stability Analysis}
Since object stability denotes the consistency of an object to have a particular pose in a scene configuration, any instabilities will decrease probability of that pose. These instabilities can be separated into translational and rotational movement, represented by a ratio between the observed object movement and the maximum movement observed when gravity is applied to the scene.

Let the object acceleration be $a_{o}$ and gravity acceleration be $g$. The translational stability of an object can be directly computed with:
\begin{equation}
\beta_{o} = \left\| \frac {a_{o}} {g} \right\|
\end{equation}

The maximum angular acceleration $\beta_{{R_{\max_o}}}$ can be computed by finding a pivot point in the object model that maximizes the torque generated by gravity. This pivot point is a point with the highest point pair distance $d_{\max}$ from object's center of gravity. Let $I$ be the object's inertia, $\alpha_{o}$ be its angular acceleration and $m$ be its mass. The maximum angular acceleration and angular stability are given by: 

\begin{equation}
\beta_{R_{\max}} = \left\| \frac{I} {g * d_{\max} * m} \right\| \text{ and } \beta_{R} = \frac{\left\|\alpha_{o} \right\|}{ \beta_{R_{\max}} }
\end{equation}

Simulation error leads to small residual movements in a scene. The stability component of the object probability must be lenient to those residual errors while penalizing other errors. This can be modeled using a logistic function given by:
\begin{resizealign} \label{eq:stability_probability}
\Pr(\mathcal{B}(o)|s) = \left[(1 + e^{-\sigma_{\mathcal{B}} (\alpha_{t} * \beta_{t} - \alpha_{\mathcal{B}})} )
(1 + e^{-\sigma_{\mathcal{B}}(\alpha_{R} * \beta_{R} - \alpha_{\mathcal{B}}) })\right]^{-1}
\end{resizealign}

\noindent
with $\alpha_t$ as the translational stability weight and $\alpha_R$ as the rotational stability weight. $\sigma_{\mathcal{B}}$ determines the steepness of the logistic function and $\alpha_{\mathcal{B}}$ determines its midpoint.
The value of $\sigma_{\mathcal{B}}$ and $\alpha_{\mathcal{B}}$ are physics engine specific. In most cases, the weight parameters, $\alpha_{t}$ and $\alpha_{R}$ can be set to $1$.

\subsection{Collision Penalty}
The presence of collisions in an estimated scene is very common, but these usually have different levels of severity.
Severity can be computed based on penetration depth and intersection volume.
Given that obtaining the exact penetration depth and intersection volume of convex objects is either an NP-hard or NP-complete problem~\cite{tiwary2008hardness,bringmann2008approximating},
it is not practical to use when evaluating a scene that may contain a large number of object pose hypotheses.
Instead, the broadphase manifold point method \cite{mirtich1997efficient} can be used for estimating penetration depth and oriented bounding box intersection can be used for estimating the intersecting volume.

The collision penalty must also be lenient to small collisions that occur due to simulation error, while being penalizing larger collisions.
Let $\Pr(\mathcal{C}(o))$ be the object collision component probability,
$\epsilon_{d}$ be the maximum distance that can be sampled from the object model,
 $d$ be the penetration depth,
  $v_c$ be the colliding volume,
  $v_{t}$ be the total object volume.
 The logistic function for the collision penalty of object $o$ is given by:
\begin{equation} \label{eq:collision_probability}
\Pr(\mathcal{C}(o)|s) = \left[1 + e^{-\sigma_c * (d / \epsilon_{d} * v_c/v_{t} - \alpha_c)}\right]^{-1}
\end{equation}

The values of $\alpha_c$ and $\sigma_c$ determine the midpoint of the probability distribution and the steepness of the logistic function respectively. Given that each physics engine might use a unique collision solver, the calculated collision severity might differ from one physics engine to the other. These values can be determined by mapping the average residuals of collision solver algorithm $e_\mathcal{C} =  d / \epsilon_{d} * v_c/v_{t}$ and then setting a desired penalty magnitude for this violation.

\subsection{Support Contribution}
The role of support contribution in scene analysis is to constraint object movements that reduce support relations between objects. This support relation is reduced when movements in the structure cause the objects supported by it to become unstable.

We employ a support probability function based on logistic regression
Let $k$ be the number of supported objects and $F_{{\mathcal{S}},i}$ describe the percentage of object $i$ weight supported by an object. The support contribution probability of object $o$ is given by:

\begin{equation}
\Pr(\mathcal{S}(o)|s) = \left[1 + e^{-\sigma_s * \sum_{i=1}^k F_{{\mathcal{S}},i} - \alpha_s}\right]^{-1}
\end{equation}
The steepness of logistic function for support contribution, $\sigma_s$, determines the strength of object probability variation when it has a different number of supported objects and $\alpha_s$ determines the function midpoint. These variables can be adjusted based on the importance of maximizing support relations in  scene estimation.

\subsection{Pose Data Fitness}
Usually the data probability $\Pr(D|s_t)$ is calculated by applying the sensor noise model over all received data points. The downside of this model is that the probability magnitude of this data fitness to the scene will scale inversely with the number of points in the scene.

In order to generate a more meaningful probability, instead of applying a noise model to all visible points in the scene, we calculate the scene data fitness based on the uncertainty of the scene model from caused by its visibility $\Pr(s_t|D)$. $\Pr(s_t|D)$ can serve as substitute because it scales linearly with $\Pr(D|s_t)$ given uniform prior for $\Pr(s_t)$ and constant $\Pr(D)$. Given that the scene model is composed of objects $O = [o_1, o_2, ..., o_N] \in s$, $\Pr(s_t|D)$ equals to uncertainty of all object poses, $\Pr(o_1, o_2, ..., o_N | D)$, caused by the limited data on the object.
The uncertainty of these objects mimics logistic regression, which is given by:
\begin{equation}
\Pr(o_i|D) = \left[1 + e^{-\sigma_d * ( c_{est}/C_{max} - \alpha_d) }\right]^{-1}
\end{equation}
The constants used in this function, $\sigma_d$ and $\alpha_d$ are set to match the accuracy profile of the selected pose estimator used when the object visibility changed.




\section{Algorithms}
Although the probability of a particular scene configuration can be computed with the scene probability components by performing the procedure described in Algorithm~\ref{alg:simulate_world}, finding the optimal scene configuration given by Eq.~\ref{eq:scene_probability_simplified} is not trivial.

Let $N$ be the number of detected objects, $k$ be the maximum number of hypotheses per object and the set of possible scene hypotheses be $S_\mathcal{H}$. Since most pose estimators do not enforce $S_\mathcal{H} \in S_C$, it is very rare for $s* \in S_\mathcal{H}$, and the optimal configuration generated from $S_\mathcal{H}$ will generally still be a bad scene model because it contains numerous physical violations.

Additionally, testing all $s \in S_\mathcal{H}$ is very inefficient with the testing complexity $O(k^N)$. 
To handle this problem, we propose a method that can find a scene new configuration with better physical consistency $s_h'$ from $s_h \in S_H$ and a strategy for approximating the optimal scene with linear complexity $O(k*N)$.


\begin{algorithm}[tb]
\caption{\strut Simulate the scene and evaluate the probability components of each object}\label{alg:simulate_world}
\begin{algorithmic} 
\Function{SimulateWorld}{simulated world $\mathcal{W}$, simulation time duration $t$, simulation step $\Delta t$, object model database $\mathcal{O}$, scene point cloud $D$, maximum point correspondence distance $d_t$, maximum generated attraction force $\mathcal{F}$}
    \State $\mathcal{W}_{-1} = $ \Call {GetPreviousSceneData}{$\mathcal{W}$}
    \For { $t_i$ in \Call{Range}{$1,t,\Delta t$}}
        \State \Call {ApplyGravity}{$\mathcal{W}$} 
        \State \Call {SolveCollisions}{$\mathcal{W}$}
        \For {$o$ in $\mathcal{W}$ }
            \State $T$ = \Call {GetObjectPose}{$o$}
            \State $i$ = \Call {GetObjectName}{$o$}
                \State $F,\tau$= \Call {GenForce}{$\mathcal{O},i,T,D,d_t,\mathcal{F}$} \Comment{Eq.~\ref{eq:total_data_force}}
                \State \Call {ApplyForcesToObject}{$o,F,\tau$}
        \EndFor 
        \State $\mathcal{G}$ = \Call {GenSupportGraph}{$\mathcal{W}$} \Comment{Alg.~\ref{alg:generate_object_support_graph}}
        \State $\mathcal{W} \leftarrow \mathcal{G}$ \Comment{Save the support graph data in world}
        \State \Call {StepSimulation}{$\Delta t$}
    \EndFor
    \State \Return \Call {SceneCondProbability}{$\mathcal{W}, \mathcal{W}_{-1},\mathcal{O},D$} \Comment{Eq.~\ref{eq:scene_probability_base}}
\EndFunction
\end{algorithmic}
\end{algorithm}

\subsection{Generating Better Scene Configurations}
We generate a new scene hypothesis $s_h'$ with better physical consistency by simulating $s_h$ in a physics engine for a short period of time. This will allow the physics engine to fix collisions and resolve stability issues caused by small estimation errors by moving the object with gravity.
Throughout, we apply guiding forces to maintain the object consistency with the sensor data.
These forces are computed based on the distance between the scene data and the current object surface points and then normalized to a certain maximum value ($\mathcal{F}$) to maintain balance with gravity.

These forces are normalized by assigning a unit vector as the maximum force vector of each point pair, such that the accumulated unnormalized guiding forces magnitude applied to the object is equal to the number of model surface points.
Let $d_t$ be the maximum correspondence distance between each point-pair, $p_{di}$ be the point coordinate in the scene with object model point correspondence $p_{ci}$, the point-pair force vector generated by this pair, $\vec{f}_{pi}$, is given by:
\begin{resizealign} \label{eq:point_data_force}
\vec{f}_{pi} =
    \begin{cases}
    (p_{di} - p_{ci}) * \frac{d_t - ||p_{di} - p_{ci}||}{d_t^2 / 4} & 0 \le ||p_{di} - p_{ci}|| \le d_t \\
    0 & \text{otherwise}
    \end{cases}
\end{resizealign}

This function is chosen because it features a smoother force magnitude transition in $d_t$ compared to a spring force model. 


These forces are combined into a centralized force and torque that is applied to the object. Let $p_{cog}$ be the object's center of gravity, $N$ be the number of point correspondences, and $M$ be the number of surface points in the model. The centralized forces $\vec{F}$ and torque $\vec{\tau}$ applied to the object is given by:

\begin{resizealign}\label{eq:total_data_force}
\vec{F} = \mathcal{F} *\frac{1}{M} \sum_{i=1}^N f_{pi}  \text{ and }
\vec{\tau} = \mathcal{F}  * \frac{1}{M} \sum_{i=1}^N (p_{ci} - p_{cog}) \times f_{pi}
\end{resizealign}

The scene after simulation $s_h'$ will have better physical consistency than $s_h$ when the object poses are reasonably close to the real scene. This means the optimal $s \in S_H'$, where $S_H'$ is the set of improved scene hypotheses, will better capture both similarity to data and physical consistency than the optimal $s \in S_H$.

\subsection{Reducing Optimal Scene Approximation Complexity}




\begin{algorithm}[tb]
\vskip 0.25cm
\caption{\strut Generate an object support graph from a simulated world}\label{alg:generate_object_support_graph}
\begin{algorithmic} 
\Function{GenSupportGraph}{simulated world $\mathcal{W}$}
    \State $\mathcal{G} = \emptyset$ \Comment A new empty support graph
    \For { colliding object pair $(o_a, o_b)$ in $\mathcal{W}$}
        \State $d_{a} = 0$; $d_{b} = 0$
        \State $vol_{a} = 0$; $vol_{b} = 0$
        \State $\vec{n}_b = 0$
        \For { contact point $p \in (o_a,o_b)$}
            \State $d_{a}$ += \Call {GetPenetrationDistance}{$p$}
            \State $d_{b}$ += \Call {GetPenetrationDistance}{$p$}
            \State $v_{a}$ += \Call {GetPenetrationVolume}{$p$}
            \State $v_{b}$ += \Call {GetPenetrationVolume}{$p$}
            \State $\vec{n}_b$ += \Call {GetNormalForces}{$p$}
        \EndFor
        \If {$\vec{n}_b < 0$} \Comment $o_b$ supports $o_a$ 
            \State $o_{p} = o_b; d_p = d_b; v_p = v_b$
            \State $o_{c} = o_a; d_c = d_a; v_c = v_a$
        \Else  \Comment $o_a$ supports $o_b$
            \State $o_{p} = o_a; d_p = d_a; v_p = v_a$
            \State $o_{c} = o_b; d_c = d_b; v_c = v_b$
        \EndIf
        \State $\mathcal{G} \leftarrow (o_{p}, o_{c})$ \Comment{Add a new edge}
        \State \Call {SetCollisionProperty}{$\mathcal{G}, o_{p}, d_{p}, v_{p}$}
        \State \Call {SetCollisionProperty}{$\mathcal{G}, o_{c},  d_{c}, v_{c}$}
    \EndFor
    \State \Return $\mathcal{G}$
\EndFunction
\end{algorithmic}
\end{algorithm}

 
The goal of evaluating scenes in $S_H$ is to maximize $\Pr(C|s_t)$.
However, a naive over all $s \in S_H$ is wasteful, because many objects in $S_H$ do not have any significant interactions with each other. These insignificant interactions happen because as long as the support relationships between objects preserved after their movements, on most cases the supported object stability will also be conserved as well.

Therefore, assuming that collisions can be resolved by the physics engine, the optimal scene configuration can be effectively found by incrementally optimizing the scene model from the base of the structure to the top of the structure.
When objects are incrementally added, we test them to ensure conservation of the support relations and object stability.



\begin{figure}[tb]
\centering
\includegraphics[width=\columnwidth]{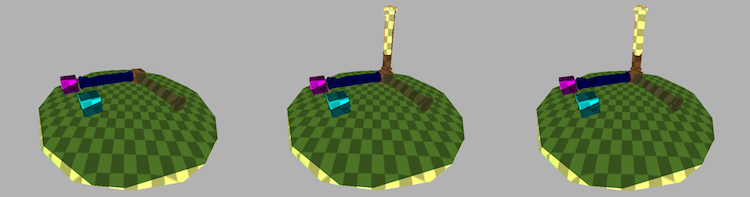}
\caption \strut
{Sample order of optimization. The left figure shows the first simulated hypotheses, where all objects are connected to the ground. The middle figure shows the supported object stability test for objects that are supporting another object. The right figure shows the best scene hypothesis result}
\label{fig:order_of_optimization}
\end{figure}

Let $V$ be the vertices of scene graph $\mathcal{G}$, which is composed of $[V_1,V_2,...]$ which represents group of vertex with increasing vertex distance to the background object and $V_i$ be a collection of vertex/object $[v_1,v_2,v_3,...]$ with equal vertex distance to the ground plane. The scene evaluation process is described in Alg.~\ref{alg:hypotheses_evaluation}, and visualized in Fig.~\ref{fig:order_of_optimization}. By using this strategy, the optimal scene estimation can be obtained with just $O(N*k)$ complexity.

\begin{algorithm}[tb]
\caption{\strut Hypotheses Evaluation}\label{alg:hypotheses_evaluation}
\begin{algorithmic} 
\Function{EvaluateHypotheses}{support graph $\mathcal{G}$, ground plane $o_b$, all objects hypotheses $\mathcal{H}$, simulated world $\mathcal{W}$, simulation time duration $t$, simulation step $\Delta t$, object model database $\mathcal{O}$, scene point cloud $D$, maximum point correspondence distance $d_t$, maximum generated attraction force $\mathcal{F}$}
    \State \Call{RemoveObjects}{$\mathcal{W}$}
    \State $V$ = \Call{VerticesByDistance}{$\mathcal{G},o_b$}
    \State $\Pr(s^*|O,D) = 0$; $s^* = \emptyset$
    \For { $V_i$ in $V$}
        \State \Call{AddObjectsInVertices}{$\mathcal{W},V_i$}
        \For {$v_j$ in $V_i$}
            \State id = \Call {GetObjectId}{$v_j$}; $H = \mathcal{H}[\text{id}]$
            \For {$o$ in $H$}
                \State $\mathcal{W}_{\text{test}} = \mathcal{W}$
                \State \Call {SetObjectPose}{id,$o, \mathcal{W}_{\text{test}}$}
                \If {\Call{HasSupportedObject}{$o$}}
                    \State \Call {AddChildOf}{$o,\mathcal{W}_{\text{test}}$}
                \EndIf
                \State params = $\{\mathcal{W}_{\text{test}},t,\Delta t, \mathcal{O} ,D,d_t,\mathcal{F}\}$
                \State $\Pr(s|O,D)$ = \Call {SimulateWorld}{params}
                \If { $\Pr(s|O,D) > \Pr(s*|O,D)$ }
                    \State $\Pr(s*|O,D) = \Pr(s|O,D)$
                    \State $\mathcal{W} = \mathcal{W}_{\text{test}}$; $s^* = \mathcal{W}$
                \EndIf
            \EndFor
        \EndFor
    \EndFor
    \State \Return $s^*, \Pr(s*|O,D)$
\EndFunction
\end{algorithmic}
\end{algorithm}

\subsection{Sequential Scene Analysis}
\begin{figure}[tb]
\centering
\includegraphics[width=\columnwidth]{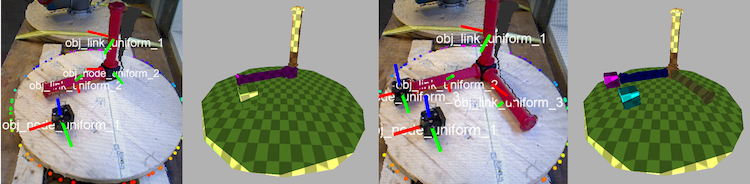}
\caption
{ An example of support retained object.}
\label{fig:support_retain_ability}
\end{figure}

There are five modes of scene transition for objects~\cite{hager_wegbreit_2011}:
\begin{enumerate}
\item Object Added: The object is not present in the previous scene but present in the current scene.
\item Object Removed: The object is present in the previous scene but not in the current scene.
\item Object Perturbed: The object is present in both previous and current scene but has moved above the predetermined threshold.
\item Object Static/Stable: The object is present in both previous and current scene with movements less than a predetermined threshold.
\item Object Support Retained: The object is missing in the current scene, but is retained because there exist other objects that are supported by this object. An example can found on Fig.~\ref{fig:support_retain_ability}.
\end{enumerate}

The first four modes of can be obtained by comparing the existence of each object in the support graph before and after the scene transition. If the object is retained, the position before and after the transition is compared to check whether the object has moved considerably or not. 

Even though the pose estimator could not detect an object, this does not mean it has been removed: it might just be occluded. Before we remove an object, we check:
\begin{enumerate}
\item If this object is still visible, move it to the static object list.
\item If this object previously supported another object and any of those objects become unstable because they do not have sufficient support, move this object to the support retained object list.
\item If this object is not visible due to occlusion and no other object seems to occupy its volume, move it to static object list and mark it as ``Ephemeral''.
\end{enumerate}
``Ephemeral'' objects will be added to the support retained list when a newly detected object requires support from this object and will be removed when a newly detected object occupies the same volume as this object.
If all these checks were false, we can remove the object from the scene.

Since the scene transition is incremental,
the good pose configurations from the previous scene can be used to estimate the current scene as additional pose hypotheses. 
Adding these previous configurations as hypotheses are beneficial because it generally has a better pose accuracy compared to hypotheses from the pose generator when the object is static and its visibility is reduced. These previous configurations also allow the scene parser to test alternate pose configurations for objects that are not visible anymore and find the best pose for those occluded object that maximizes the overall scene probability given the object hypotheses and sensor data.

\subsection{Software Implementation}
\begin{figure}[tb]
\centering
\includegraphics[width=\columnwidth]{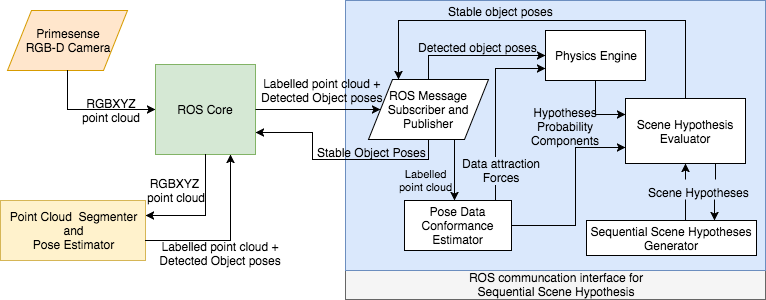}
\caption{System Data Flow Diagram}
\label{fig:figure_system_data_flow_diagram}
\end{figure}

Our software used Boost for multithreading, Point Cloud Library (PCL) for manipulating scene data, ObjRecRANSAC \cite{papazov_haddadin_parusel_krieger_burschka_2012} for generating object hypothesis, BulletPhysics for simulating the scene and computing the scene probability, and the Robot Operating System (ROS) for communication between Primesense RGB-D camera, the point cloud segmentation pipeline, and our software. The diagram in figure~\ref{fig:figure_system_data_flow_diagram} describes the flow of data in our system.
\section{Experiments}
We evaluated the pose fluctuation and accuracy of our SSP algorithm on both simulated and real point cloud data.
The simulated point cloud data is generated with Gazebo 2.2.3 using plain cubes, plain blocks, and scanned textured meshes of hand tools.
The real point cloud data was captured with a Primesense Carmine 1.09 RGB-D camera.
For each example, the scene parsing algorithm was performed 4-5 times to test the scene consistency on static scene before adding more objects in the scene to test the scene update ability.

\subsection{Simulated Data}

In the simulated data experiments, we compared the error between the best object configuration in each frame and the ground truth poses. The pose hypotheses for each object were generated by adding noise to the ground truth pose which increases as the visibility of an object drops. We assess precision of pose estimates as the magnitude of pose fluctuations in a the static scene.

\begin{figure}[tb]
\centering
\includegraphics[width=\columnwidth]{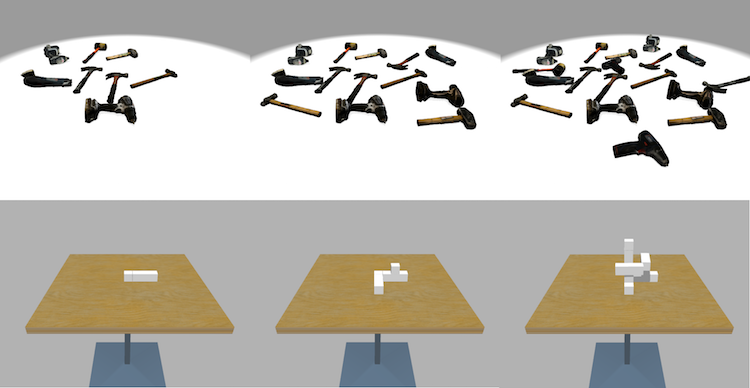}
\makeatletter
\let\@currsize\normalsize
\caption{Sample object configurations in the isolated and complex structure object experiment}
\label{fig:simulated_object_test}
\end{figure}

We performed experiments on two sets of objects. The first experiment observes the scene parsing performance for isolated objects, which are composed of various mallets, drills, and a sander. Objects in the scene do not interact with each other. The purpose of this experiment is to clarify the performance of the scene parsing on individual objects. The experiment was done using 8 different tool models and up to 15 objects was present for each frame. Figure~\ref{fig:simulated_object_test}(top) shows captured color images of some frames used in this experiment.

The second experiment assesses performance of the SSP algorithm when building a complex structure with simulated blocks and cubes.
Figure~\ref{fig:simulated_object_test}(bottom) shows the RGB data from each frame. When some objects are completely obstructed by the other object, the scene parsing algorithm will use its previous scene knowledge to estimate these obstructed object pose.


Pose accuracy was defined by the estimation error to the ground truth pose, and precision was defined by the distance between an estimated pose to the average estimated pose of an object. The accuracy and precision of the both isolated and complex structure test can be found on Table~\ref{tab:simulated_isolated_complex_result}.

\begin{table}[tb]
\centering
\caption{Isolated and Complex Object Test Result}
\label{tab:simulated_isolated_complex_result}
\begin{tabular}{llll}
\Xhline{2\arrayrulewidth}
\multicolumn{1}{c}{Test} & \multicolumn{1}{c}{Category} & \multicolumn{1}{c}{\begin{tabular}[c]{@{}c@{}}Translation\\ Errors (mm)\end{tabular}} & \multicolumn{1}{c}{\begin{tabular}[c]{@{}c@{}}Rotation\\  Errors (deg)\end{tabular}} \\ \hline
\multirow{3}{*}{\begin{tabular}[c]{@{}l@{}}Isolated\\ Object\end{tabular}}  & Induced Noise & 3.05 $\pm$ 1.11 & 3.35 $\pm$ 1.2 \\
 & Scene Parsing Accuracy & 5.25 $\pm$ 0.96 & \textbf{2.79 $\pm$ 2.92} \\
 & Scene Parsing Precision & 1.57 $\pm$ 1.27 & 0.72 $\pm$ 1.2 \\ \hline
\multirow{3}{*}{\begin{tabular}[c]{@{}l@{}}Complex\\ Structure\end{tabular}} & Induced Noise & 5.19 $\pm$ 2.81 & 5.7 $\pm$ 3.02 \\
 & Scene Parsing Accuracy & \textbf{3.65 $\pm$ 3.41} & \textbf{2.23 $\pm$ 1.94} \\
 & Scene Parsing Precision & 2.49 $\pm$ 3.34 & 1.23 $\pm$ 1.25 \\ 
\Xhline{2\arrayrulewidth}
\end{tabular}
\end{table}

\subsection{Real Data}
The objects used in this experiment were a black magnetic block and a red magnetic beam, which were placed in a stable configuration. These objects have self-aligning property when connected to each other, which makes the assembled object pair have a consistent relative pose.

The ground truth object pose could not be observed directly, but the accuracy of estimation can be measured indirectly by observing the relative transform between objects in contact with one another due to this self-aligning property. The relative transform is measured between all connected block and beam pairs and compared with the known relative transform between each object pair. 



\begin{table}[tb]
\centering
\caption{Accuracy and Precision Test Result}
\label{tab:accuracy_precision_test_result}
\resizebox{\columnwidth}{!}{%
\begin{tabular}{llll}
\Xhline{2\arrayrulewidth}
Test & \multicolumn{1}{c}{Category} & \multicolumn{1}{c}{\begin{tabular}[c]{@{}c@{}}Translation\\ Errors (mm)\end{tabular}} & \multicolumn{1}{c}{\begin{tabular}[c]{@{}c@{}}Rotation \\ Errors (deg)\end{tabular}} \\ \hline
\multirow{2}{*}{Accuracy} & ObjRecRANSAC & 11.28 $\pm$ 5.42 & 19.00 $\pm$ 6.42 \\
 & Sequential Scene Parsing & \textbf{6.92 $\pm$ 4.12} & \textbf{11.81 $\pm$ 5.62} \\ \hline
\multirow{2}{*}{Precision} & ObjRecRANSAC & \textbf{2.05 $\pm$ 2.09} & 5.376 $\pm$ 6.66 \\
 & Sequential Scene Parsing & 3.75 $\pm$ 2.89 & \textbf{3.866 $\pm$ 3.33} \\ 
\Xhline{2\arrayrulewidth}
\end{tabular}%
}
\end{table}

\begin{figure}[tb]
\centering
\includegraphics[width=\columnwidth]{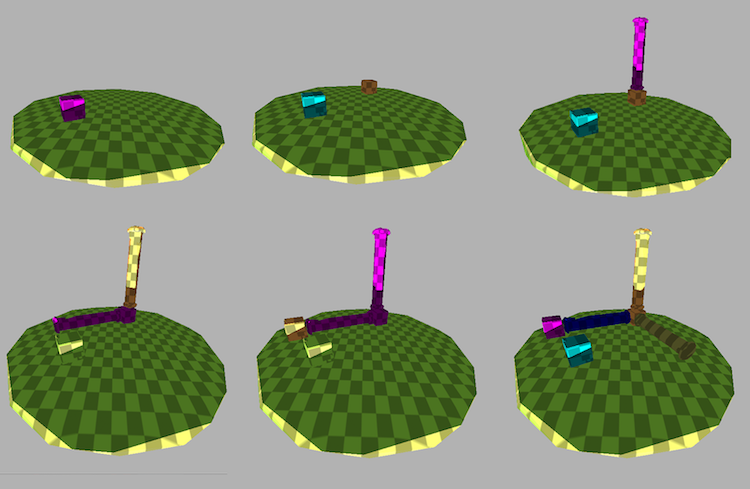}
\makeatletter
\let\@currsize\normalsize
\caption{Incremental changes of scene in real data test.}
\label{fig:increasing_scene_complexity}
\end{figure}

The precision of the scene estimation was measured by recording the fluctuation of the best scene estimation poses of each object for each newly detected objects. The result of the accuracy and precision test can be found on Table~\ref{tab:accuracy_precision_test_result}, while a sample sequence of simulated world generated when performing this experiment can be found in Figure~\ref{fig:increasing_scene_complexity}.



\section{Case Study}
\begin{figure}[tb]
\centering
\includegraphics[width=\columnwidth]{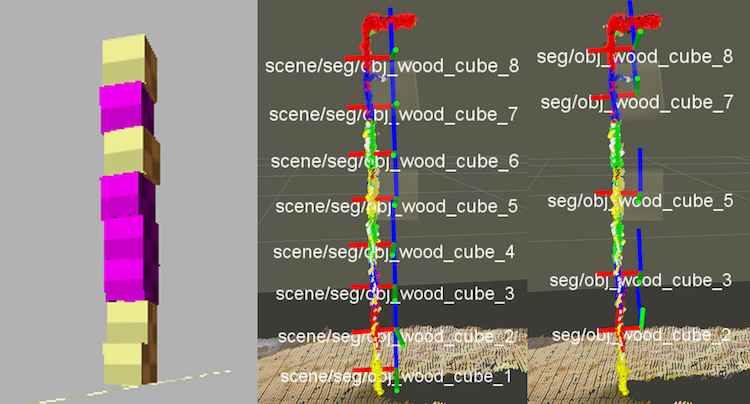}
\begin{tabular}{c c c}
Scene Parsing World & Scene Parsing & ObjRecRANSAC
\end{tabular}
\caption{\strut Target Assembly Structure detected by both ObjRecRANSAC and Scene Parsing. The scene parsing pose have an offset in the gravity direction, because the colored blocks shape and dimension do not perfectly match its model}
\label{fig:robot_final_structure}
\end{figure}

Finally, we performed a case study of a simple assembly task. It was performed with a UR5 robot with Robotiq 2-finger gripper controlled using the CoSTAR system~\cite{paxton2017costar}. The target structure consists of a tower that contains as many colored blocks as possible, picked from a set of 8 2-inch colored blocks placed at random on a table. The final structure is shown on Figure~\ref{fig:robot_final_structure}, while environment setup used in this case study can be viewed in Figure \ref{fig:robot_initial_setup}.


The task was repeated 5 times for both our SSP algorithm and for the pose estimation algorithm ObjRecRANSAC~\cite{papazov_burschka_2011} alone.
The performance was evaluated based on the straightness of the structure from the base, and we measured consistency of the detected blocks.
The straightness of the structure is measured by the horizontal component of the distance (x-y plane) between the base of the structure to each block in the structure.
Since this distance will accumulate as the number of blocks increased, it was normalized by dividing each measurement with the height of the tower to the observed block. 
We quantify pose consistency as the fluctuation of poses over time. The results of this experiment are shown in Table~\ref{tab:case_study_result}.


\begin{table}[tb]
\centering
\caption{Case Study Result}
\label{tab:case_study_result}
\begin{tabular}{llll}
\Xhline{2\arrayrulewidth}
\multicolumn{2}{l}{Category} & ObjRecRANSAC & SSP \\ \hline
\multicolumn{2}{l}{Successful Stacks} & 5 $\pm$ 1.58 & \textbf{6.2 $\pm$ 0.45} \\ \hline
\multirow{2}{*}{\begin{tabular}[c]{@{}l@{}}Assembly\\ Straightness\end{tabular}} & Translation (mm) & 1.72 $\pm$ 1.7 & \textbf{1.3 $\pm$ 1.25} \\
 & Rotation (deg) & 2.61 $\pm$ 2.83 & \textbf{0.79 $\pm$ 1.2} \\ \hline
\multirow{2}{*}{\begin{tabular}[c]{@{}l@{}}Pose\\ Consistency\end{tabular}} & Translation (mm) & \textbf{3.8 $\pm$ 4} & 5.59 $\pm$ 9.14 \\
 & Rotation (deg) & 14.57 $\pm$ 9.18 & \textbf{6.7 $\pm$ 8.65} \\ \hline
\Xhline{2\arrayrulewidth}
\end{tabular}
\end{table}

\section{Discussion}
The experiments on simulated scenes, real scenes, and object stacking show that the proposed SSP algorithm is consistently better at estimating object orientations than noisy input hypotheses. 

However, SSP could only fix small object translation errors. 
When evaluating SSP, all object poses are tested for both accuracy and precision, including objects with very low visibility which were prone to having more errors, while ObjRecRANSAC pose evaluation can only be done for an object with reasonably high visibility and thus had a bias towards less error.

Furthermore, since the physics engine used in our algorithm (Bullet3) used a simplified collision mesh generated from a convex decomposition process \cite{mamou2009simple}, the inconsistency between the collision model and the real object geometry may introduce additional noise. This bias can be seen on Fig.~\ref{fig:robot_final_structure}, where the height of block does not match with the actual block position because the blocks used in the case study had varying shapes and sizes.

Still, the SSP algorithm was highly successful and has more consistent performance in the block stacking test compared to ObjRecRANSAC alone. This is mainly attributed to the ability of SSP to fix orientation errors, which enables more consistent block grasping and constrained the orientation of the block placement so the manipulated block surface to aligned with the top of the structure. 

Further, while a direct comparison was impossible, experimental results achieve better pose accuracy at longer distances than the previous system proposed by Brucker at al.~\cite{brucker_leonard_bodenmuller_hager_2012}. 
Our experiments have generally been done 1 meters away from the RGB-D sensor, and reports 7 mm average translation error 12 degrees orientation error for blocks and beams in experiments including some object occlusions. Experimental results in \cite{brucker_leonard_bodenmuller_hager_2012} showed between 1-2 cm translation errors and 10-20 degrees orientation errors for objects without occlusions.

\section{Conclusions}
We described an algorithm to perform Sequential Scene Parsing, which incrementally builds a scene model from sensor data using a physics engine and information from the previous scene. This scene model is generated by finding the optimal scene configuration that maximizes a probability function that enforces data, physical and temporal consistencies of the scene.
These probability functions are divided into several categories; each imposes different constraints and parameters that are necessary for the generation of physically consistent scene estimates.
These functions have several variables that can be tuned to prioritize one constraint over the other constraint, which is capable of shaping the optimal scene result. We also proposed an efficient scene optimization strategy that approximates the best scene with linear complexity. To validate our approach, we performed object pose accuracy and precision evaluations on both simulated and real data with various object models.

In the future, we plan to support input scene models generated with SLAM-based techniques such as KinectFusion and add support for more complex object interactions such as magnetic and grasping interactions. This would enable more accurate scene approximation and improved the applicability of our solution to more objects. 

\addtolength{\textheight}{-12cm}   





\bibliographystyle{IEEEtran}
\bibliography{references}

\end{document}